\documentclass[lettersize,journal]{IEEEtran}
\usepackage{amsmath,amsfonts}
\usepackage{algorithmic}
\usepackage{algorithm}
\usepackage{array}
\usepackage[caption=false,font=normalsize,labelfont=sf,textfont=sf]{subfig}
\usepackage{textcomp}
\usepackage{stfloats}
\usepackage{url}
\usepackage{verbatim}
\usepackage{graphicx}
\usepackage{cite}
\usepackage{soul,color} 
\usepackage{hyperref} 
\usepackage{booktabs} 

\begin{document}

\title{In-orbit demonstration of object-based change detection applied to the maritime domain}


\author{
    Cristopher Castro Traba$^{1,*}$, Tom Hendrix$^1$, Amaury Perrocheau$^1$, Francesco Pasanisi$^1$, Finian Rogers$^1$, Aubrey Dunne$^1$, David Rijlaarsdam$^1$, Roberto Del Prete$^2$, Nicolas Long{\'e}p{\'e}$^2$, Gabriele Meoni$^{2,3}$
    
    \thanks{$^1$Ubotica Technologies, Kluyverweg 1, Delft, 2629 HS, South Holland, Netherlands}
    
    \thanks{$^2$$\Phi$-lab, European Space Research Institute (ESRIN), European Space Agency (ESA), Via Galileo Galilei, 1, Frascati, 00044, RM, Italy}

    \thanks{$^3$Advanced Concepts and Studies Office,  European Space Research Institute (ESRIN), European Space
Agency (ESA), Via Galileo Galilei, 1, Frascati, 00044, RM, Italy}
    
    \thanks{Cristopher Castro Traba is the corresponding author: cristopher.traba@ubotica.com}
}


\markboth{Preprint. Under review.}%
{Castro Traba \MakeLowercase{\textit{et al.}}: In-orbit change detection demonstration}

\maketitle

\begin{abstract}

This article presents the successful in-orbit validation of an object-based change detection pipeline for low latency maritime monitoring onboard the CogniSAT-6 mission. We propose and validate a lightweight, three-phase operational sequence: optimized ground-based dual-acquisition scheduling, real-time onboard AI-based vessel segmentation, and multi-temporal matching. Upon acquiring consecutive images over a targeted Area of Interest (AoI), the onboard system correlates vessel detections using a hybrid matching approach that integrates the Hungarian algorithm for optimal assignment with RANSAC-driven spatial and size filters for outlier rejection. A series of in-orbit validations demonstrated robust performance across 11 diverse maritime scenarios, achieving valid recall and precision values (considering only vessel size and location) exceeding 0.8. The validation demonstrated the ability of the system to track both cooperative targets and uncooperative "dark" vessels strictly through geometric consistency. This capacity to autonomously generate and act upon real-time intelligence establishes a foundation for next-generation Insight-as-a-Service (IaaS) and tipping-and-cueing architectures. 

\end{abstract}


\section{Introduction} \label{s: introduction}

\IEEEPARstart{T}{raditional} Earth Observation (EO) operations rely predominantly on a bent-pipe paradigm where mission plans are uploaded as static sequences to be executed. In this model, satellites act as passive data acquisition nodes, capturing and transmitting sensor data to ground stations for subsequent processing\cite{rijlaarsdam2024next}. By the time an image is analyzed on the ground, hours or even days may have passed, generating the insights with a latency too high to be useful for time-sensitive maritime operations \cite{amaro2020}.

This latency limitation is critical in the maritime domain. The surveillance of the global waters is a fundamental component of international security, environmental protection, and economic stability. EO platforms equipped with optical, Radio-Frequency (RF) and Synthetic Aperture Radar (SAR) payloads are making a significant contribution  in supporting the authorities in their efforts. These platforms provide global coverage to detect, classify, and track marine vessels to combat illegal, unreported, and unregulated (IUU) fishing, prevent maritime piracy, detect illegal ship-to-ship trans-shipments and maritime air and sea pollution, and disrupt smuggling operations \cite{amaro2020}. Furthermore, detailed monitoring of vessel traffic allows for the optimization of port logistics, maritime threat assessment, accident investigation, future vessel behavior prediction, and the estimation of the spatial distribution and intensity of fishing activities \cite{zissis2016real}. Because the time-scale of vessel movements demands real-time intelligence, the maritime domain serves as an ideal operational proving ground for next-generation satellite autonomy.

The shift toward autonomous remote sensing was pioneered by the Autonomous Sciencecraft Experiment (ASE) onboard the Earth Observing-One (EO-1) spacecraft, which proved that satellites could intelligently react to their environment \cite{chien2004eo}. Through onboard re-scheduling, ASE demonstrated the ability to modify mission schedules in real-time based on observed data and internal spacecraft constraints. This evolution from execution to goal-oriented autonomy has been further catalyzed by recent missions, such as the $\Phi$-Sat-1, $\Phi$-Sat-2 and CogniSAT-6 missions \cite{rijlaarsdam2024next,giuffrida2021varphi}. Concepts validated in the aforementioned missions pave the way for more elaborate applications, such as detecting changes in the extracted data between different acquisitions taken by the same satellite at distinct times.


This article presents a novel object-based change detection pipeline, validated via on-orbit experiments, proving the feasibility of object based change detection onboard a satellite. Concepts explored in recent works have been focused on in-orbit change detection of full-scale raw imagery, mainly for downlink efficiency \cite{serief2023deep, guerrisi2022satellite}. This work proves the feasibility of extracting insights based on acquisition-to-acquisition changes, potentially using them to drive a change in internal operational schedule, unlocking satellite autonomy.

Instead of comparing full-scale raw imagery, the proposed system operates via an object-detection pipeline:

\begin{itemize}
    \item Acquisition and Segmentation: An initial image is acquired over an Area of Interest (AoI), an onboard vessel segmentation algorithm is executed, and the vessels' spatial attributes are saved to an onboard database.
    \item Subsequent Pass: Upon a subsequent orbital pass over the same AoI, a new image is acquired and processed through the same segmentation model.
    \item Point Cloud Matching: An onboard matching algorithm compares the new detections against the historical database using only geographical coordinates and measured object dimensions to correlate vessels across acquisitions.
\end{itemize}

The architecture of this technological demonstration is not limited to a single satellite but designed for scalability.
The operational utility of satellite constellations is constrained by the high latency of bent-pipe architectures. Mission value can be increased by transitioning constellations toward autonomous closed-loop constellations. This requires individual satellites to evaluate acquisition-to-acquisition changes using onboard edge processing to perform internal schedule adjustments or cross-platform tasking \cite{zeleke2023new, thangavel2023trusted}. This type of capabilities enables future constellations to operate as a "system-of-systems," where decentralized coordination is facilitated by Inter-Satellite Links (ISL) \cite{zilberstein2025federated}.

The successful demonstration of on-orbit change detection unlocks numerous use cases. Extracted changes can be used to autonomously trigger a third acquisition without ground intervention, enabling the continuous tracking of IUU fishing fleets or vessels encroaching on critical infrastructure. In the energy sector, this capability provides insights for real-time commodities monitoring \cite{arslanalp2019big}, where the movement of Very Large Crude Carriers (VLCCs) from anchorages to offloading sites can be instantly detected and transmitted via ISL, as well as enabling automatic monitoring of vessel anchorages, choke point queues or ports. 

The main contributions of this article are as follows: 
\begin{itemize}
    \item Design and implementation of an object-based change detection solution onboard a spacecraft.
    \item In-orbit demonstration of the matching solution under multiple environmental conditions in a maritime monitoring application.
    \item Early initiation of a road map towards increased autonomy of (federated) constellations, enabling more elaborate autonomous tipping and cueing.
\end{itemize}

The remainder of this article has the following structure. Section \ref{s: related work} describes the related work in maritime monitoring, on-orbit processing and current solutions for change detection in EO applications, and commercial value of the proposed work. Section \ref{s: methodology} presents the proposed multi-temporal change detection solution for vessel monitoring.  
Section \ref{s: results} presents the results of the on-ground and in-orbit verification of the matching solution with multiple environmental conditions in CogniSAT-6. Section \ref{s: discussion} analyses the results and discusses further applications of the proposed solutions, and Section \ref{s: conclusion} concludes the study.

\section{Related work} \label{s: related work}

\subsection{Maritime Domain Awareness and Tracking}
Maritime Domain Awareness (MDA) via Earth Observation is a well-established discipline. Vessel segmentation has been heavily validated, accelerated in part by open-source initiatives such as the Kaggle Airbus Ship Detection Challenge \cite{multipleApproachesAirbus}. Consequently, a multitude of segmentation architectures have been developed, such as Neural Networks (NN) applied to optical and Synthetic Aperture Radar (SAR) satellite imagery \cite{NearRealTimeSAR}. Operationally, these remote sensing techniques are often fused with Automatic Identification System (AIS) data to track fleets. However, AIS tracking is critically limited by "dark" vessels that disable their transponders to evade detection. This limitation underscores the need for independent, vision-based tracking for effective maritime traffic monitoring, fisheries management, and maritime conservation \cite{matchingNeed}.

\subsection{In-Orbit Processing and Autonomy}

Achieving near-real-time intelligence for dynamic EO use cases without relying on continuous downlinks requires a shift toward autonomous onboard edge computing. Massive raw images can be analyzed directly onboard to immediately downlink only lightweight insights, a paradigm that simultaneously bypasses traditional communication bottlenecks and maximizes end-user value \cite{paravano2023value}. Recent advances in spacecraft edge computing have demonstrated the feasibility of these concepts \cite{phisatOnboardAIApps, rijlaarsdam2024next, opssat}. Despite this progress, deploying computationally heavy AI-based models in orbit remains severely constrained by the hardware limitations of state-of-the-art satellites \cite{barretta2026towards}. Consequently, while static onboard object detection is maturing \cite{del2025optimizing,traba2026towards}, complex multi-temporal analysis on the edge remains a largely unaddressed challenge. The efficiency of satellite operations can be further improved by improving the autonomy of the individual satellite and/or any constellation that it is part of \cite{araguz2018applying}.

\subsection{The Gap in Multi-Temporal Moving Target Analysis}

Bridging this gap in autonomy requires the ability to evaluate dynamic environments across sequential satellite passes.
Multi-temporal analysis has seen considerable research for tracking slow-evolving phenomena, such as glacier dynamics or urban expansion \cite{glacier,urbanExpansion}. However, applying multi-temporal analysis to fast-moving targets from space is traditionally reliant on heavy, ground-based processing. For example, single-acquisition techniques, such as wake analysis, can estimate a vessel's speed and heading \cite{del2025robust}, but only acquisition-to-acquisition evaluation can confirm and measure the spatial displacement. 

The continuous monitoring of targets across temporal satellite acquisitions has traditionally relied on pixel-based change detection, a methodology that is often computationally expensive for hardware-constrained onboard environments. Pixel-based change detection algorithms operate by comparing the radiometric values of individual, corresponding pixels across two co-registered images. This approach requires both the Day 1 and Day 2 images to be present in the satellite's active memory simultaneously, imposing a considerable memory burden on the spacecraft. Additionally, pixel-based change detection is highly sensitive to co-registration errors. If the two high-resolution images are not perfectly aligned, it will result in massive amounts of false positives along the scene \cite{chen2012object}. In maritime monitoring specifically, transient environmental factors such as sun glint and atmospheric interference are frequently detected as radiometric changes, further degrading the reliability of pixel-level comparisons. 

To solve these processing bottlenecks and environmental sensitivities, object-based change detection methods prioritize extracted features over raw pixels, offering a more resilient and resource-efficient framework for onboard monitoring \cite{walter2004object}. Considering only extracted insights rather than the entire images reduces the processing data volume from gigabytes to kilobytes. Furthermore, focusing on detections reduces the need for exact georeferencing of the two acquisitions, as the algorithm focuses on discrete centroid correlations rather than aligned pixel comparisons. In other domains such as aviation, recent frameworks using satellite imagery have demonstrated the tracking of moving airplanes over non-radar zones \cite{planess}. Similarly, remote sensing is increasingly used to track heavy logistical assets in terrestrial environments \cite{humaninducedchanges}. However, the current literature in these domains relies mostly on computationally demanding machine learning techniques not suitable for mobile or edge deployment \cite{planess}. The object-based matching proposed offers a low-latency onboard alternative for object tracking, specifically demonstrated for the use case of maritime monitoring for vessel tracking.

\section{Methodology} \label{s: methodology}

\begin{figure}
    \centering
    \includegraphics[width=0.8\linewidth]{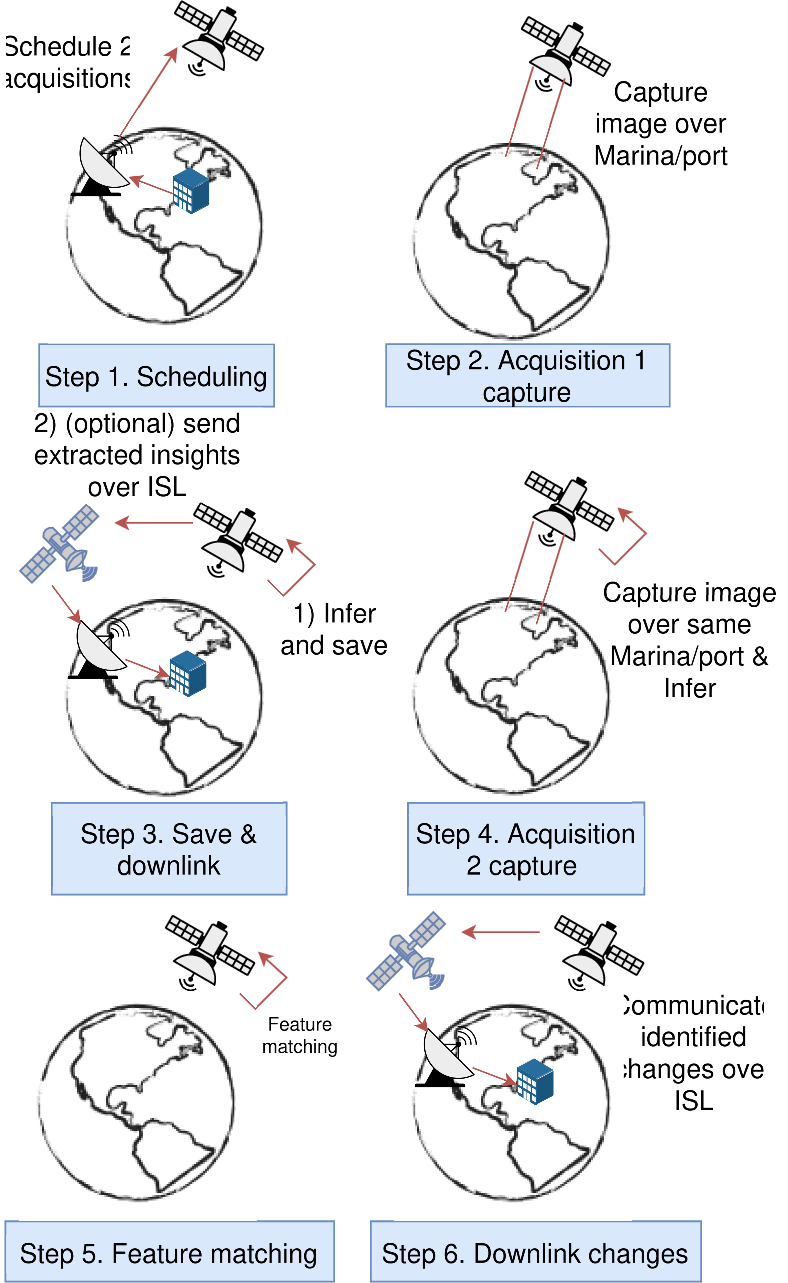}
    \caption{Diagram for the multi-temporal on orbit change detection solution.}
    \label{fig:matchingOverview}
\end{figure}

The proposed change detection framework is deployed on the CogniSAT-6 mission, leveraging its advanced onboard processing capabilities \cite{rijlaarsdam2024next}. The methodology is divided into three distinct phases: acquisition scheduling, onboard vessel segmentation, and the multi-temporal change detection algorithm. The main operational steps of the change detection framework are illustrated in \autoref{fig:matchingOverview}.

\subsection{Acquisition Strategy and Re-tasking} 

The data acquisition strategy relies on scheduling consecutive observations in a specific Area of Interest (AoI) to establish a temporal baseline. To achieve this, the scheduler generates a dual-request for the same target coordinates, dynamically adjusting the satellite's attitude to minimize the total revisit time. Varying roll-angles conditions per acquisition introduces spatial offsets and footprint variations between the two images. Consequently, the matching algorithm is designed to consider only the detections within the geographical intersection (common area) of the two observations.

The target AoIs were selected to evaluate the solution across diverse operational scenarios with variable vessel distribution. AoIs characterized by relatively sparse vessel distribution include the ports of Long Beach (USA), Piraeus (Greece), and Dakar (Senegal). On the other hand, Port Said (Egypt), and Khor Fakkan (UAE) were chosen to evaluate the performance of the solution in high-density port areas, and Gibraltar was included to assess the performance in areas with predominant moving vessels. These three categories provide a diverse test set to validate the performance of the matching solution under real-world maritime conditions.

\subsection{Onboard Real-Time Vessel Segmentation}

Upon image acquisition, the onboard pipeline executes a vessel segmentation model using the CogniSAT-XE2 platform \cite{rijlaarsdam2024next}. This model runs inference on raw imagery to minimize the end-to-end latency between image acquisition and insight generation. The resulting segmentation mask is immediately processed to extract the geographical location, physical dimensions and orientation of detected vessels. These detections serve a dual purpose: they are downlinked immediately via Inter-Satellite Link (ISL) as Real-Time Insight Delivery (RTID), and simultaneously stored in an onboard database. In the first acquisition, all the detections are downlinked through ISL, while on the second acquisition only the unmatched vessels are downlinked.

The geographical coordinates of the detections are achieved via a parametric mapping between the image plane and the geographical coordinate system. To minimize computational load onboard, the geometric coefficients defining this mapping are derived during the ground tasking phase, based on the expected AoI, and uploaded to the spacecraft. The onboard postprocessing algorithm performs a parametric interpolation that dynamically compensates for the orientation and scaling variations induced by the projected footprint, allowing for the precise derivation of latitude and longitude, as well as width and height for the pixel coordinates.

\subsection{Multi-Stage Change Detection Algorithm}

The core of the change detection methodology is a multi-stage matching algorithm designed to correlate vessel detections across temporal acquisitions. To generate these matches, the current implementation relies exclusively on the geographical location and vessel dimensions extracted during postprocessing. 

Prior to the matching algorithm there are two sequential filters to reduce the number of vessels considered for matching. First, a dimensional filter discards targets below a configurable length threshold, focusing the analysis on vessels of interest. Second, a spatial consistency filter removes detections located outside the overlapping region of the two images using a Ray Casting algorithm. Since the footprints of consecutive passes do not perfectly align due to variations in orbital trajectory and sensor attitude (particularly roll angle), this step eliminates false positives arising from the non-overlapping AoIs. 

\begin{figure}
    \centering
    \includegraphics[width=0.8\linewidth]{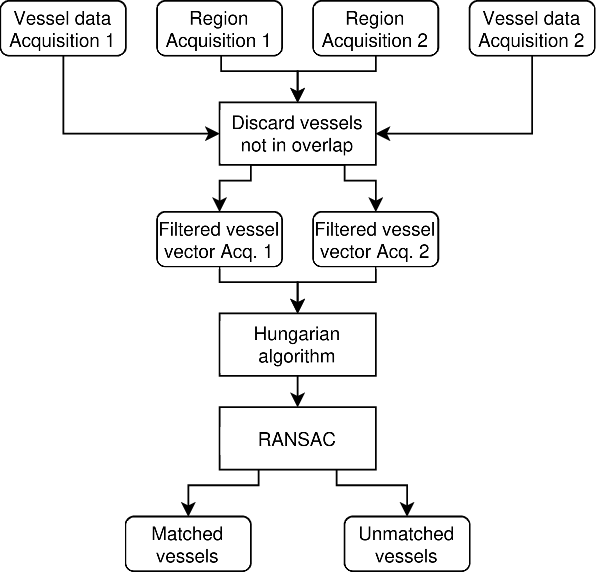}
    \caption{Block diagram depicting the on board matching flow.}
    \label{fig:matchingalgorithm}
\end{figure}

Following spatial filtering, the solution employs a hybrid approach combining the Hungarian algorithm for assignment and Random Sample Consensus (RANSAC) for robust outlier rejection, with an approximate total complexity of $\mathcal{O}(n^{3} + kn)$ \cite{munkres1957algorithms,fischler1981random}. Where $n$ is the number of rows of the cost matrix used by the Hungarian algorithm and $k$ is the number of RANSAC iterations. Finally, a validation step checks the size deviation of the matching pairs, and retains as valid matches only those exhibiting a size difference of less than a defined difference from the mean value. 
The complete diagram for this multi-step matching process is depicted in Figure \ref{fig:matchingalgorithm}, and the following subsections describe the different components and the validation methods.

\subsubsection{Common Area filter}

\begin{figure}
    \centering
    \includegraphics[width=0.9\linewidth]{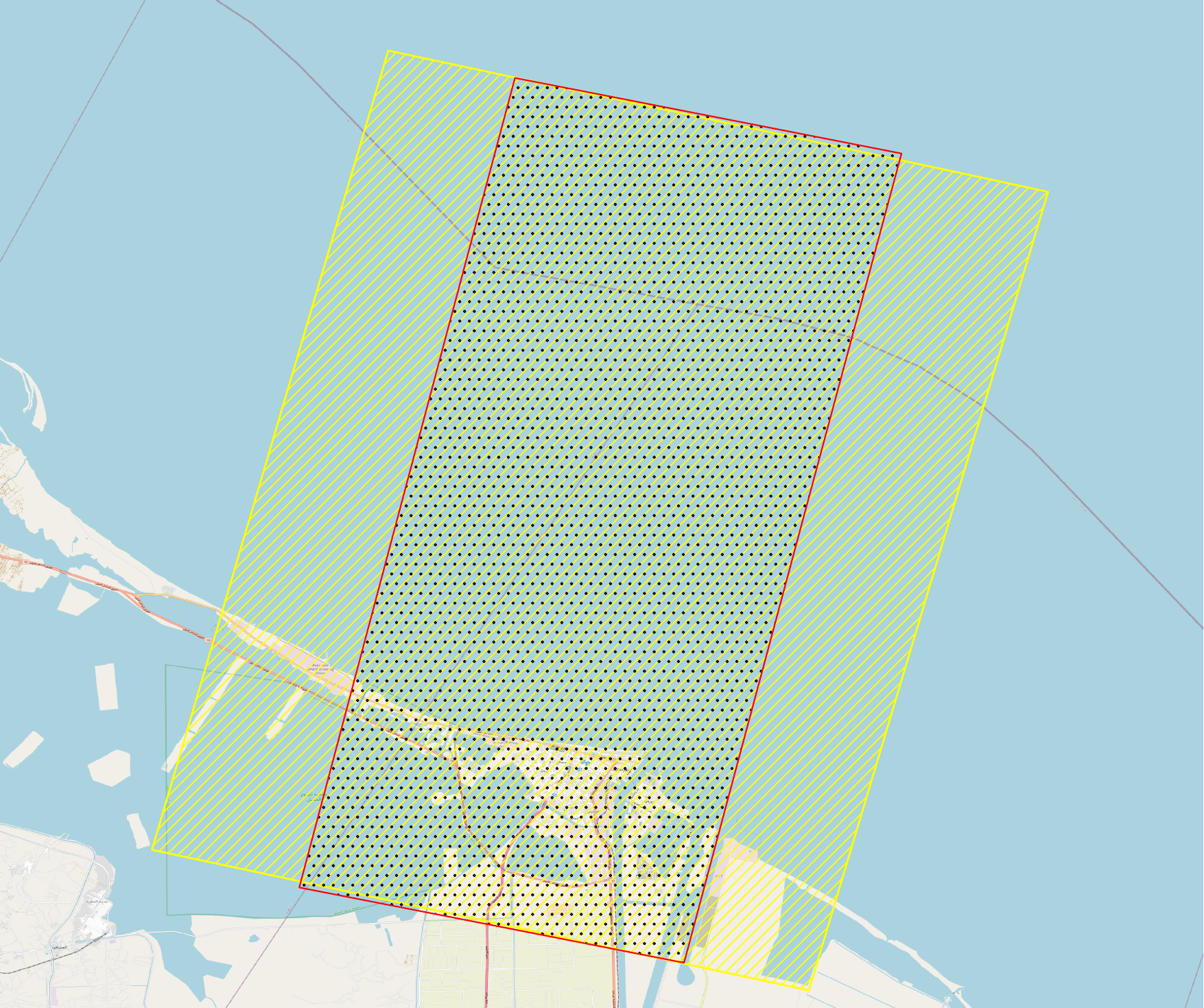}
    \caption{Visual depiction of the area difference covered by the satellite under different roll angle acquisitions. In dotted red the acquisition with small roll angle and in yellow striped the acquisition with larger roll angle.}
    \label{fig:geojsonoverlap}
\end{figure}

The Ray Casting Algorithm is used to determine if the detected points are within the intersection area of the acquisitions \cite{roth1982ray}. This method checks that the points are within the vertical limits of the intersection area, and if a horizontal line extending to the right from the point crosses a border an odd number of times, the point is classified as interior to the AoI. \autoref{fig:geojsonoverlap} illustrates an example of the roll angle effect, showing that one acquisition doubles the area covered by the other. For this reason the spatial filter is applied, removing all those detections outside the common area of both acquisitions.

\subsubsection{Hungarian algorithm}

The core component of the matching solution is the Hungarian Algorithm (Kuhn-Munkres), which addresses the vessel association problem by minimizing the global assignment cost between point sets \cite{munkres1957algorithms}. This optimization method inputs a cost matrix and determines the optimal set of row-column pairs, ensuring that the total cost or dissimilarity between the detections of both acquisitions is minimized.

Prior to optimization, a cost matrix $C$ to minimize of size $m\times n $ ($n \geq m$) is constructed, where $m$ and $n$ represent the number of detections in the first and second acquisitions. Each element $C_{i,j}$ quantifies the dissimilarity between detection $i$ from the first acquisition and detection $j$ from the second acquisition. The cost function is a normalized, weighted combination of spatial distance and vessel size similarity. Spatial distance $d_{i,j}$ is computed using the Haversine formula to account for Earth's curvature, while size costs are derived from the relative differences in vessel length ($h$) and width ($w$). The cost for each pair is calculated according to \autoref{eq: cost matrix cost calculation}:

\begin{equation} \label{eq: cost matrix cost calculation}
    C_{i,j} = \dfrac{d_{i,j}}{d_{max}} + \dfrac{|h_{i} - h_{j}|}{\frac{h_{i} + h_{j}}{2}\cdot \delta_{h}} + \dfrac{|w_{i} - w_{j}|}{\frac{w_{i} + w_{j}}{2}\cdot \delta_{w}} 
\end{equation}

where $d_{max}$ represents the maximum allowable distance, and $\delta_{h}$ and $\delta_{w}$ represent the allowable deviation factors in a 0-1 scale for height and width, respectively. To enforce geometric constraints, pairs exceeding $d_{max}$ or the size difference are assigned an infinite cost to effectively prohibit their association. These variables are defined as the coarse matching distance and the coarse size thresholds used when defining the validation conditions, and they are determined empirically.

The Hungarian algorithm provides an optimal assignment based on detection-pair costs, but it does not ensure geometric displacement consistency for all the matches. To address this, the matches identified with the Hungarian algorithm are refined using a simplified RANSAC. This step aims to estimate the global mean displacement vector between the two acquisitions and to eliminate matches that do not conform to this translation model. The RANSAC implementation considers the same spatial and size constraints as the cost matrix used in the Hungarian algorithm. A confidence based on the spatial alignment error $E_{dist,k}$ is derived by computing the mean residual Haversine distance (in meters) for each RANSAC inlier $k$ after projecting the matched coordinates from the second acquisition back to the first using the optimal computed transformation. Its formula is depicted in \autoref{eq: Confidence RANSAC}.

\begin{equation} \label{eq: Confidence RANSAC}
    C_{RANSAC,k} = \exp\left( - \frac{E_{dist,k}^2}{\sigma_{dist,k}^2} \right)
\end{equation}

Finally, to reduce false positives, a strict post-filtering stage is applied to the matches identified by RANSAC. This filter rejects any match where the length or width deviation between the detections in both acquisitions exceeds a specified value. This step is needed for eliminating false positives where objects may be spatially aligned but morphologically different. The individual length and width errors, $E_{L,k}$ and $E_{W,k}$, are subsequently aggregated into a single size confidence score $C_{size} \in (0, 1]$. This score is computed via a Gaussian decay function designed to strictly penalize average dimensional deviations that surpass the predefined tolerance threshold, $\sigma_{size}$. Its formula is depicted in \autoref{eq: size confidence}.

\begin{equation} \label{eq: size confidence}
    C_{size,k} = \exp\left( - \frac{(\frac{E_{L,k} + E_{W,k}}{2})^2}{\sigma_{size}^2} \right)
\end{equation}

Finally, the overall combined confidence score, $C_{total,k}$, is derived as the arithmetic mean of the size confidence and the geometric RANSAC confidence. Its formula is depicted in \autoref{eq: total confidence}.

\begin{equation} \label{eq: total confidence}
    C_{total,k} = \frac{C_{size,k} + C_{RANSAC,k}}{2}
\end{equation}

\subsubsection{Model outputs}

In the final RTID message, only those unmatched vessels will be downlinked. Matched vessels are static over the images, so they are not objects of interest for this application. This optimizes the communication with ground by reducing the total amount of data to be transmitted. This is particularly interesting in CubeSats and for ISL, where the communication bandwidth is limited.

\subsubsection{Metrics and validation tests definition}

The performance of the matching solution was verified on ground across 14 images (7 pairs) under multiple environmental conditions and has been validated in-orbit across 6 images (3 image pairs) on the CogniSAT-6 spacecraft. The target AoIs include diverse operational scenarios with variable vessel distribution: Long Beach (USA), Piraeus (Greece), Dakar (Senegal), Port Said (Egypt), Khor Fakkan (UAE), and Gibraltar. The performance was evaluated on the ground using reference data derived from Automatic Identification System (AIS). The validation framework distinguishes between two categories of correct matches: Identity Matches ('Real') and Feasibility Matches ('Valid'). An Identity Match is confirmed when the paired detections correspond to the same unique AIS vessel identifier in both acquisitions. On the other hand, a Feasibility Match occurs when the detections correspond to different AIS vessel identifiers that present similar dimensions and location within the matching defined tolerance thresholds. Accounting for Feasibility Matches is critical, as the onboard matching solution only has access to size and location information, not the vessel identity.

Performance is measured using Precision and Recall. Precision is defined as the ratio of correct matches, both Identity and Feasibility, to the total number of matches generated by the onboard solution. Recall represents the system's detection rate, calculated as the ratio of correct onboard matches to the total number of valid matches existing in the ground truth. Both metrics are evaluated separately for the Identity and Feasibility matches. 

The complete matching algorithm is subjected to seven variables: vessel length threshold, coarse matching distance, coarse size deviation, final matching distance, final size deviation, number of RANSAC iterations, and minimum matching confidence. The vessel length threshold is the minimum vessel length required by a detection to be considered for matching. The coarse matching distance and coarse size deviation are the values used in the cost matrix computation to limit the number of points considered for matching. These are set larger than the final matching distance and size deviation used in the final filter, to generate a broad initial set of potential matching candidates and further refine them at the end. The final matching distance and final size deviation are the maximum allowable distance and size differences to consider a theoretically correct match, and the minimum matching confidence is the minimum value to consider a theoretically correct match based on the average sum of the RANSAC and morphological confidences.  The values, empirically determined, used in the in-orbit validations are presented in \autoref{tab: Matching parameters values for the in-orbit validation}.

\begin{table}
    \caption{Matching parameters values for the in-orbit validation}
    \label{tab: Matching parameters values for the in-orbit validation}
    \centering
    \begin{tabular}{ll}
    \toprule

        \textbf{Parameter} & \textbf{Value [units]}\\ \midrule
        Vessel length threshold & 60 [m]\\
        Final matching distance & 1,000 [m]\\
        Coarse matching distance & 20,000 [m]\\
        Final size difference & 20 [\%] \\
        Coarse size difference & 30 [\%] \\
        Number of RANSAC iterations & 2000 [-]\\
        Minimum matching confidence & 50 [\%] \\
        \bottomrule
    \end{tabular}
\end{table}

\section{Results} \label{s: results}

Table \ref{tab: Matching results in CogniSAT-6 operational runs} presents the validation results of the proposed matching solution on the CogniSAT-6 satellite. The results contain data from two distinct operational campaigns: an initial on-ground verification campaign utilizing 14 images (7 image pairs) comprising 999 vessel data points, and a subsequent on-orbit validation on 6 images (3 image pairs) comprising 242 vessel data points extracted from raw imagery processed directly onboard.

\begin{figure*}
    \centering
    \includegraphics[width=\linewidth]{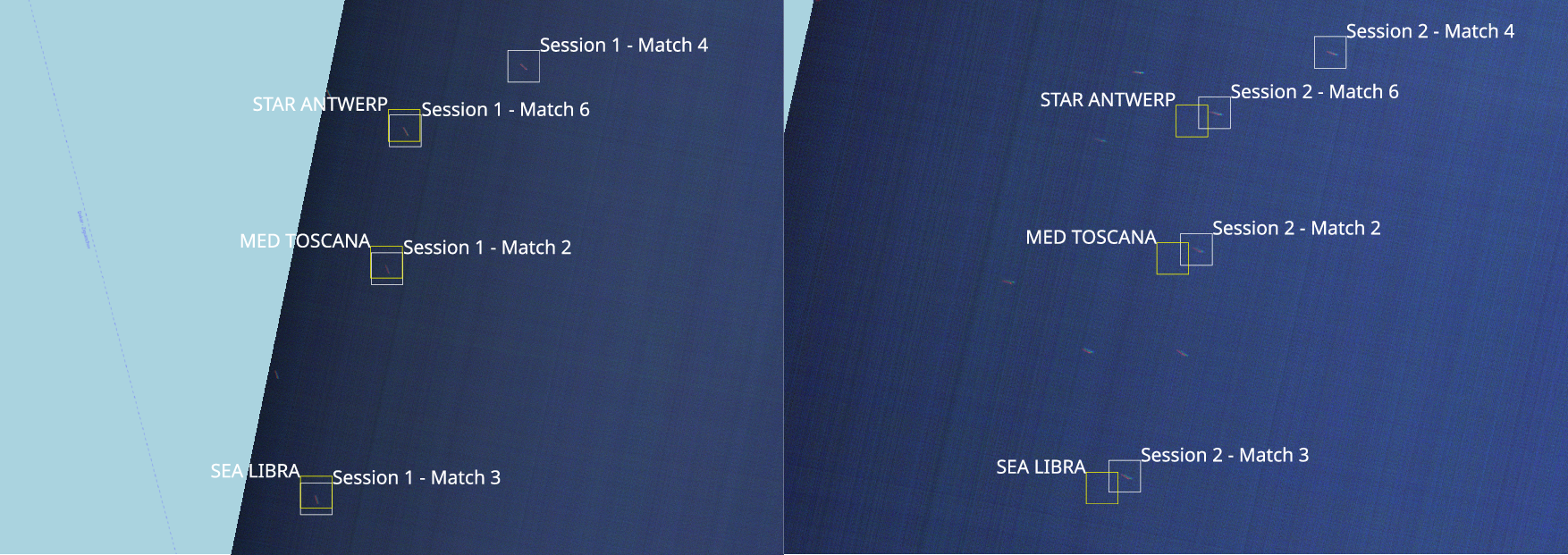}
    \caption{Matching performance under low vessel density conditions in the Dakar images taken on the 28/12/2025 (left) and the 20/12/2025 (right). The white boxes correspond to the identified matches considering the minimum vessel length, while the yellow boxes correspond to the AIS matches with the same criteria. Match 5 within the acquisitions is a dark vessel (without AIS).}
    \label{fig: Matching performance Dakar.}
\end{figure*}

\begin{figure*}
    \centering
    \includegraphics[width=\linewidth]{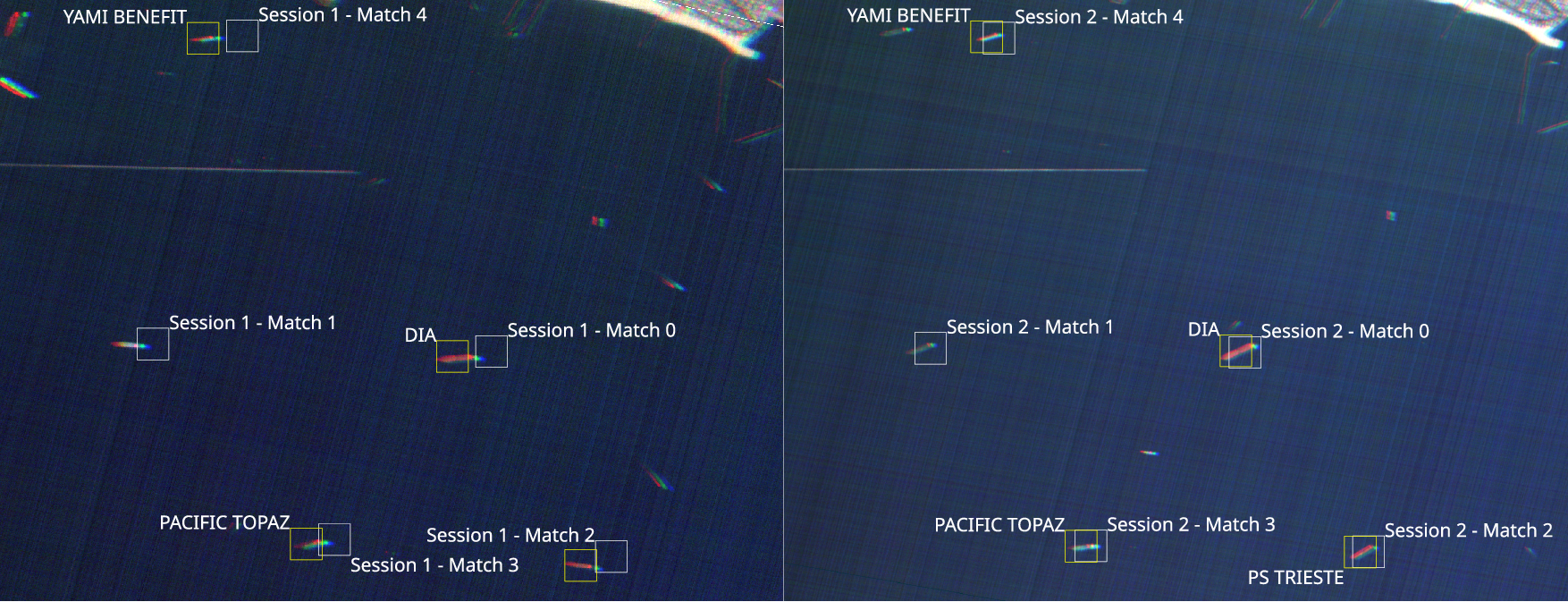}
    \caption{Matching performance under low-medium vessel density in the Long Beach images taken on the 13/09/2025 (left) and the 16/09/2025 (right). The white boxes correspond to the identified matches considering the minimum vessel length, while the yellow boxes correspond to the AIS matches with the same criteria.}
    \label{fig: Matching performance Long Beach first example correct.}
\end{figure*}

\begin{figure*}
    \centering
    \includegraphics[width=\linewidth]{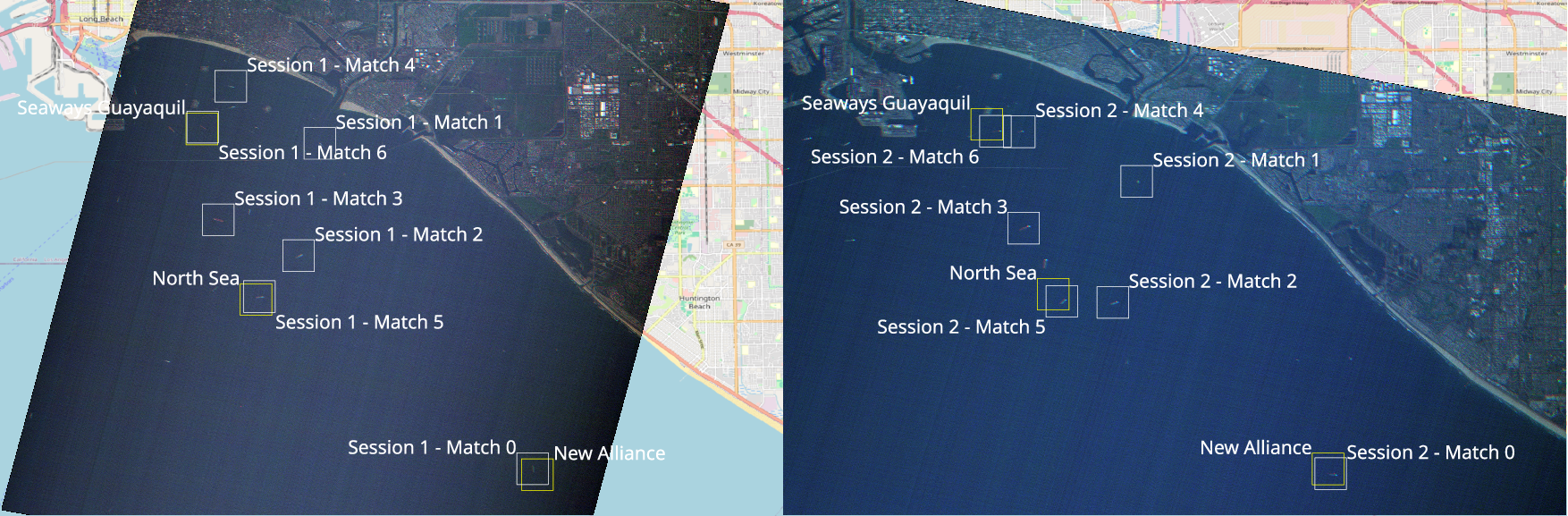}
    \caption{Matching performance under low-medium vessel density conditions in the Long Beach images taken on the 10/01/2026 (left) and the 12/01/2026 (right). The white boxes correspond to the identified matches considering the minimum vessel length, while the yellow boxes correspond to the AIS matches with the same criteria.}
    \label{fig: Matching performance Long Beach second example correct.}
\end{figure*}

As shown in \autoref{tab: Matching results in CogniSAT-6 operational runs}, the proposed matching solution achieved a consistent high detection capability over diverse environmental scenarios. Both Valid Recall and Valid Precision exceeded 0.80 for all but one acquisition pair. The system demonstrated robust performance in low-to-medium density environments (5-30 vessels), as can be seen in the Long Beach acquisitions (\autoref{fig: Matching performance Long Beach first example correct.} and \autoref{fig: Matching performance Long Beach second example correct.}), where the solution achieved a Valid Precision of 1.0. In these images, the system successfully detected all vessel matches with verified AIS, while the matches without AIS corresponded to vessels of identical type, dimensions, and location. This explains why the Real Recall and Real Precision present lower values, particularly in high-density locations, where similar vessels occupy the same anchorage areas. Furthermore, this solution demonstrated a robust capability for detecting and tracking uncooperative "dark" vessels across temporal acquisitions. In the Dakar images depicted in \autoref{fig: Matching performance Dakar.}, the matching solution detected 8 out of 10 verified AIS matches while successfully re-detecting a dark vessel across both acquisitions entirely through geometric feature matching. 
In all comparison figures, the left image represents the first acquisition and the right image represents the second. White bounding boxes highlight algorithmic matches that satisfy the minimum vessel length constraint, whereas yellow bounding boxes indicate AIS-verified matches under the same constraint. In some cases, the yellow boxes are not perfectly aligned with the vessel due to the georeferencing accuracy of the image. Additionally, each detection includes a text label to facilitate visual traceability between the two images.

\begin{table*}[h]
\centering
\begin{minipage}{\textwidth}
\centering
\caption{Matching results in CogniSAT-6 operational runs.}
\label{tab: Matching results in CogniSAT-6 operational runs}
\begin{tabular}{lcccc}
 \toprule
    \noalign{\vskip 0.3em}
\multicolumn{1}{c}{Matching pair run} &
\multicolumn{1}{c}{Valid Recall} &
\multicolumn{1}{c}{Valid Precision} &
\multicolumn{1}{c}{Real Recall} &
\multicolumn{1}{c}{Real Precision} 
  \\[0.3em] \midrule
    \noalign{\vskip 0.3em}

Long Beach (13/09/25-16/09/25) & 0.8333 & 1.0000 & 1.0000 & 0.8000 \\

Port Said (17/11/25-20/11/25)& 0.9706 & 0.9167 & 0.3810 & 0.4211 \\

Khor Fakkan (30/10/25-01/11/25) & 0.8750 & 0.8400 & - & - \\

Gibraltar (04/12/25-07/12/25)* & 1.0000 & 1.0000 & - & - \\

Piraeus (10/12/25-12/12/25)& 1.0000 & 1.0000 & 0.3333 & 0.3333 \\

Dakar (28/12/25-30/12/25)& 0.8889 & 1.0000 & 0.8000 & 0.8890 \\

Long Beach (10/01/26-12/01/26) & 1.0000 & 1.0000 & 0.7500 & 0.7500\\

Long Beach (24/01/26-26/01/26) & 0.7500 & 1.0000 & 0.5000 & 0.3333\\

Dakar (15/02/26-17/02/26)* & 1.0000 & 1.0000 & 0.6666 & 1.0000\\

Port Said (26/02/26-28/02/26)* & 1.0000 & 1.0000 & 0.3000  & 0.8182 \\

\bottomrule
\end{tabular}

\vspace{5pt} 
    \footnotesize 
    \textit{Note:} * corresponds to the runs validated in orbit, and - in the Real Recall and Real Precision corresponds to runs where no AIS was obtained due to the acquisition conditions.
\end{minipage}

\end{table*}

\section{Discussion} \label{s: discussion}

Ground profiling and in-orbit demonstrations confirm the viability of multi-temporal, object-based change detection for maritime monitoring.  This discussion explores the implications of these results, the operational boundaries of the system, and its potential to shape the future of autonomous marine monitoring.

The experiments demonstrated that onboard vessel matching using spatial position and vessel size is highly robust across diverse operational scenarios. In moderately trafficked waters, relying on vessel position data alone risks false matches. By including vessel size, the algorithm successfully filters out false positives while maintaining a computational overhead low enough to be viable for edge-computed environments. The performance of the solution was consistent across acquisitions. However, an analysis of specific image pairs highlights the system's sensitivity to environmental factors, vessel density variations, and preprocessing inaccuracies. Specifically, the Real Recall and Real Precision (dependent on AIS-verified matches) are considerably lower than the Valid metrics because vessels of the same type stay within the same anchorage locations. 

\begin{figure*}
    \centering
    \includegraphics[width=\linewidth]{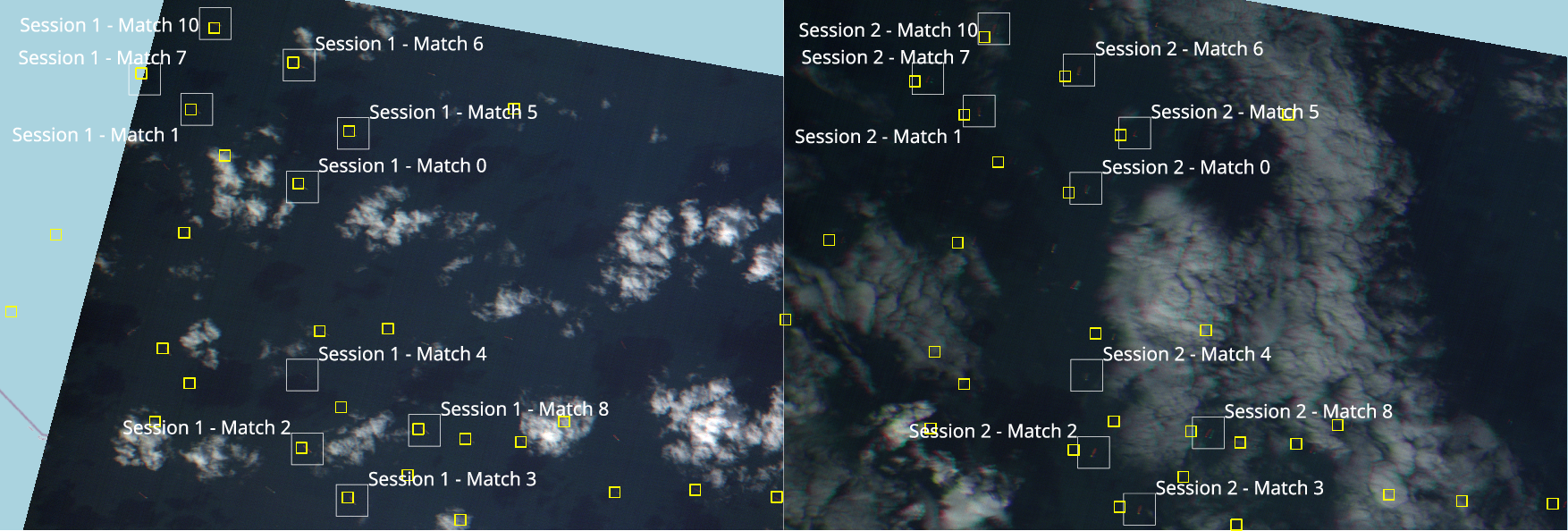}
    \caption{Matching performance under high vessel density and cloudy conditions in the Port Said images taken on the 26/02/2026 (left) and the 28/02/2026 (right). The white boxes correspond to the identified matches, while the yellow boxes correspond to the AIS matches.}
    \label{fig: Matching performance under high vessel density and cloudy conditions in the Port Said images taken on the 26/02/2026-28/02/2026.}
\end{figure*}

\begin{figure*}
    \centering
    \includegraphics[width=\linewidth]{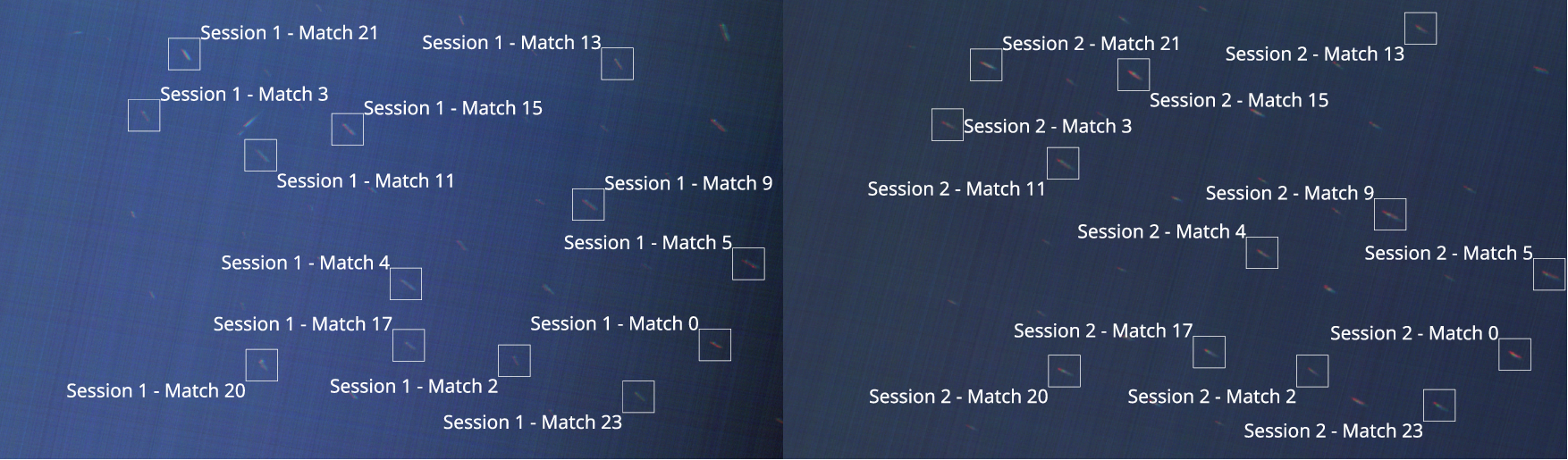}
    \caption{Matching performance under high vessel density conditions in the Khor Fakkan images taken on the 30/10/2025 (left) and the 01/11/2025 (right). The white boxes correspond to the identified matches.}
    \label{fig: Matching performance Khor Fakkan.}
\end{figure*}

\begin{itemize}
    \item \textbf{Dense and Uncooperative Environments:} Dense maritime environments (more than 50 vessels) present unique validation challenges due to the large number of similar ships within the same area. For example, in the Khor Fakkan acquisitions taken on 30/10/2025-01/11/2025, the presence of numerous "dark" vessels (ships not broadcasting AIS) prevented definitive ground-truth validation, despite the algorithm successfully pairing them, as shown in \autoref{fig: Matching performance Khor Fakkan.}. 
    A similar validation performance occurred in the Port Said images taken on 17/11/2025-20/11/2025, where high density combined with georeferencing inaccuracies over an open-sea anchorage highlighted the challenge of distinguishing real operational matches from likely false positives. These two cases showcase why the Real metrics (AIS-verified) are lower than the Valid metrics. Despite these validation limitations, the Dakar results (\autoref{fig: Matching performance Dakar.}) confirm that the proposed geometric feature correlation is capable of tracking dark vessels.

    \item \textbf{Impact of Meteorological Conditions:} Cloud cover remains a primary operational constraint in optical payloads. Cloudy conditions reduce the total number of detections within the image and decrease the maximum achievable Real Recall rate because some of the AIS-verified matches are covered by clouds. In the Port Said images taken on 26/02/2026-28/02/2026 (\autoref{fig: Matching performance under high vessel density and cloudy conditions in the Port Said images taken on the 26/02/2026-28/02/2026.}), the majority of AIS-verified matches were completely covered by clouds. Despite the cloudy conditions and high vessel density, the matching solution demonstrated resilience, matching 9 out of 30 real matches, with 9 of its 11 total matches verified as real with AIS. Similarly, the Piraeus acquisitions taken on 10/12/2025-12/12/2025 suffered from heavy cloud cover in the second image, limiting the system to detecting only 1 out of 3 AIS-verified matches.

    \item \textbf{Sensitivity to Georeferencing Accuracy:} Beyond environmental factors, accurate image georeferencing has a significant effect on the matching algorithm, as spatial alignment errors affect the intersection logic in the matching solution. In the Port Said images taken on 26/02/2026-28/02/2026, poor geolocation between the image pair led to an incorrectly computed intersection area, initially resulting in a Recall and Precision of 0. The values reported in \autoref{tab: Matching results in CogniSAT-6 operational runs} and illustrated in \autoref{fig: Matching performance under high vessel density and cloudy conditions in the Port Said images taken on the 26/02/2026-28/02/2026.} for this Port Said pair were obtained by removing the GeoJSON intersection filter, clearly highlighting the algorithm's vulnerability to pointing errors. These cases highlight the onboard matching solution's reliability is tied to the georeference accuracy of the acquisitions.

\end{itemize}

Downlink requirements are reduced from gigabytes of raw imagery to mere bytes of telemetry through onboard processing. This transition from a "data downlink" to an "insight downlink" cuts the response latency from hours to minutes, since only the coordinates of the detected objects need to be transmitted. This also mitigates tasking latency, since on-orbit extraction and matching allow satellites to autonomously re-task over an area of interest, capturing moving targets that traditional bend-pipe architectures would otherwise miss. Integrating this matching solution with onboard AIS receivers will allow future payloads to instantly cross-reference a vessel's physical footprint with its broadcasted identity, providing a powerful, automated safeguard against AIS spoofing.

The proposed matching solution extends far beyond maritime environments use cases. The underlying change detection architecture is sensor-agnostic, it can be applied across heterogeneous spacecraft, supporting diverse EO applications. Detecting a high-interest event change enables direct Inter-Satellite Link (ISL) communication for rapid "tip and cue" of trailing sensors in a constellation.  Applied to higher-resolution optical imagery, it can monitor for example mining excavators and terrestrial logistics. Integrated with thermal sensors, it enables predictive alerting for natural hazards, such as volcanic eruptions or active wildfire tracking.

\section{Conclusion} \label{s: conclusion}

This article demonstrated the successful in-orbit technical demonstration of a multi-temporal, object-based change detection system. It was validated onboard the CogniSAT-6 mission in 6 acquisitions (3 image pairs) under multiple environmental conditions. While verification across diverse acquisition conditions confirmed the algorithm's robustness, it also highlighted operational constraints requiring further refinement, specifically pointing accuracy, cloud cover, track ambiguities among similar vessels, and current limitations in feature extraction. The main limiting factor found regarding matching performance across the diverse operational runs was the accuracy of the geolocation between successive acquisitions. Future work includes adding visual features to the matching algorithm. Such features can be represented as embeddings and compared with an AI model to provide an additional confidence for the match found.

Despite these physical and environmental boundaries, validating this edge-computed change detection paves the way for autonomous satellite operations through internal schedule adjustments. At a foundational level, onboard confirmation of an unchanged scene enables the complete omission of unnecessary data downlinks. Direct object-based matching will drive advanced operational capabilities, including autonomous onboard re-tasking, fleet tracking, real-time anchorage utilization monitoring, and commodity tracking. It is important to note that this matching  solution is not exclusive for maritime EO applications. The algorithm processes generalized sets of insights, it is inherently adaptable to other domains and sensor modalities. 

Ultimately, a single satellite cannot provide the revisit frequency required for persistent global tracking. Therefore, the critical next step in making this technology operational is enabling distributed computation and insight-sharing across multi-satellite architectures. Because the validated matching algorithm is largely sensor-agnostic, it can inherently ingest and correlate point clouds across varying resolutions and modalities. This flexibility lays the groundwork for heterogeneous, multi-constellation networks where an insight generated by one satellite seamlessly tips and cues the next, realizing a fully integrated, autonomous EO system.

\section*{Acknowledgments}
The authors would like to thank the European Space Agency $\Phi$-lab Italy for the support provided through the FutureEO-1 block 4 project [Contract No. 4000146318/2024/I-DT].

\bibliographystyle{IEEEtran}
\bibliography{references}

\section{Biography Section}

\begin{IEEEbiographynophoto}{Cristopher Castro Traba}
eceived the Master of Science degree in space engineering with Cum Laude distinction from Delft University of Technology, Delft, Netherlands. He is currently a Space Systems Engineer for Ubotica Technologies, where he manages and designs AI-based solutions for EO missions. He has previously been a Visiting Researcher at the European Space Agency where he implemented Deep Learning models for thermal hotspot segmentation in raw multispectral imagery , an Undergraduate Researcher at the Universidade de Vigo where he developed relative pose estimation algorithms for autonomous spacecraft using LiDAR , and a Graduate Researcher at TU Delft where he researched ML for spacecraft telemetry analysis on resource-constrained microprocessors.
\end{IEEEbiographynophoto}

\begin{IEEEbiographynophoto}{Tom Hendrix}
received the Master of Science degree in aerospace engineering from the Delft University of Technology, Delft, The Netherlands, specializing in Space System Engineering and performing thesis research on radiation effects in a soft processor on an FPGA using proton irradiation. He currently works for Ubotica Technologies as part of the Space System R\&D team, where he is a Senior Space System Engineer and Flight Software Developer for several space missions.
\end{IEEEbiographynophoto}
\begin{IEEEbiographynophoto}{Amaury Perrocheau}
received a Master degree in aerospace engineering from ISAE-SUPAERO and another in Astrophysics from Paul Sabatier University. He previously led a gravitational wave observation program for the SETI Institute. He is currently a space systems engineer for Ubotica Technologies where he is involved in the daily operations of the  CogniSAT-6 satellite and is developing efficient scheduling algorithms to increase the value obtained from Earth observations. Amaury is also involved in the integration of third-party software and demonstrations such as onboard applications provided by NASA/JPL.
\end{IEEEbiographynophoto}

\begin{IEEEbiographynophoto}{Francesco Pasanisi}
received a Master of Science in Geographic Information Science and Remote Sensing from Wageningen University \& Research (WUR), and a Master of Science in Agricultural and Environmental Science from the University of Bari "Aldo Moro". He previously served as a Research and Teaching Assistant at WUR, where he developed deep learning datasets for agricultural crop segmentation and assisted in teaching Remote Sensing courses. He is currently a Geospatial Machine Learning Engineer at Ubotica Technologies, where he develops end-to-end geospatial data workflows and engineers Earth Observation solutions for the training and deployment of deep learning models across cloud environments and on-board satellite hardware. Francesco has authored multiple peer-reviewed publications on remote sensing and deep learning, and was awarded the Best Poster Award at the IEEE GRSS IADF School on Computer Vision for Earth Observation, sponsored by the European Space Agency.

\end{IEEEbiographynophoto}
\begin{IEEEbiographynophoto}{Finian Rogers}
received a Bsc of Computer Science in 1955 at the university of Westminster. He is currently senior program manager at Ubotica Technologies. Previous experience as a program manager at Intel leading multiple projects. Most notably, driving DARPA researchs programs within Intel Programmable Solutions Group.
\end{IEEEbiographynophoto}
\begin{IEEEbiographynophoto}{Aubrey Dunne}
received the B.Eng. degree in mechatronic engineering, the M.Eng. degree in electronic
systems, and the Ph.D. degree in computer vision, all from Dublin City University, Dublin, Ireland. He is currenly a Co-Founder and Chief Technology Officer (CTO) with Ubotica Technologies, a Space AI company with headquarters in Dublin, Ireland. Ubotica specializes in AI processing at the edge, deploying power-efficient embedded solutions in the Space segments. He led the Ubotica team that in 2020 successfully delivered the AI solution enabling the world’s first AI inference on board an EO CubeSat. Prior to his role in Ubotica, he gained ten years of professional engineering consultancy experience, with companies including Intel and Movidius, in the core areas of computer vision and embedded systems design.
\end{IEEEbiographynophoto}

\begin{IEEEbiographynophoto}{David Rijlaarsdam}
received the Master of Science degree in aerospace engineering from Delft University of Technology, Delft, The Netherlands, with a specialization in Space System
Engineering. He is currently the Director of Space System Engineering for Ubotica Technologies, where he manages the Space System R\&D team and is involved as a System Engineer in several space missions. He has previously been part of the Automation and Robotics Section of the European Space Agency where he researched relative navigation for spacecraft and has been part of the advanced architecture team of Intel Movidius, where he researched the application of AI for star identification algorithms.
\end{IEEEbiographynophoto}
\begin{IEEEbiographynophoto}{Roberto Del Prete}
received the M.Sc. degree (cum laude) in aerospace engineering, and the Ph.D. degree in industrial engineering disserting about “artificial intelligence onboard satellites” from the University of Naples Federico II, Naples, Italy, in 2019, 2021, and 2024, respectively. He is currently an Internal Research Fellow with $\Phi$-lab, European Space Agency (ESA), Frascati, Italy, specializing in deep learning and edge computing for remote sensing. He focuses on improving time-critical decision-making through advanced AI solutions for space missions and Earth monitoring. His notable work includes the development of “FederNet,” a terrain relative navigation system. His professional experience includes roles as a Visiting Researcher with the $\Phi$-Lab, European Space Agency and SmartSat CRC, Adelaide, SA, Australia. He has contributed to key projects like Kanyini Mission, and developed AI algorithms for real-time maritime monitoring and thermal anomaly detection. He co-authored of more than 30 scientific publications, he is dedicated to leveraging advanced technologies to address global challenges in remote sensing and AI.

Dr. Prete codeveloped the award-winning P$^{3}$ANDA project, a compact AI-powered imaging system, was the recipient of the 2024 Telespazio Technology Contest prototype prize.
\end{IEEEbiographynophoto}
\begin{IEEEbiographynophoto}{Nicolas Long{\'e}p{\'e}}
received the M.Eng. degree in electronics and communication systems and the M.Sc. degree in electronics from the National Institute for the Applied Sciences, Rennes, France, in 2005, and the Ph.D. degree in signal processing and telecommunication from the University of Rennes I, Rennes, in 2008.

From 2007 to 2010, he was with the Earth Observation Research Center, Japan Aerospace Exploration Agency, Tsukuba, Japan. From 2010 to 2020, he was a Research Engineer with the Space Observation Division, Collecte Localization Satellites, Plouzan{\'e}, France. Since September 2020, he has been an Earth observation data Scientist, $\Phi$-Lab Explore Office, European Space Research Institute, European Space Agency, Frascati, Italy. He has been working on the development of innovative synthetic aperture radar-based applications for environmental and natural resource management (ocean, mangrove, land and forest cover, soil moisture, snow cover, and permafrost) and maritime security (oil spills, sea ice, icebergs, and ship detection/tracking). At the $\Phi$-Lab, he is particularly involved in the development of innovative Earth Observation missions in which artificial intelligence is directly deployed at the edge (on the spacecraft). His research interests include Earth observation remote sensing and digital technologies, such as machine (deep) learning.
\end{IEEEbiographynophoto}

\begin{IEEEbiographynophoto}{Gabriele Meoni}
received the M.Sc. degree in electronic engineering and the Ph.D. degree in informa-
tion engineering, under the supervision of Prof. Luca Fanucci, from the University of Pisa, Pisa, Italy, in 2016 and 2020, respectively. After completing his doctoral studies, from 2020 to 2023 he was a Internal Research Fellow with the European Space Agency (ESA) (Advanced Concepts Team (ACT) 2020–2021 and $\Phi$-lab 2021–2023), where he conducted research on Artificial Intelligence (AI) and neuromorphic computing for onboard spacecraft applications. During 2022–2023, he was a visiting Researcher with AI Sweden, focusing on distributed edge learning for satellite constellations. From 2023 to 2024, he was an Assistant Professor with the Faculty of Aerospace Engineering at Delft University of Technology. He is currently an Innovation Officer with ESA, Frascati, Italy. He has coauthored more than 40 scientific publications. His research interests include satellite onboard processing, AI for Earth observation, and neuromorphic computing\end{IEEEbiographynophoto}

\end{document}